\documentclass[lettersize,journal]{IEEEtran}
\usepackage{amssymb}
\usepackage{hyperref}
\usepackage{booktabs}
\usepackage{multirow}
\usepackage{xcolor}
\usepackage{graphicx}
\usepackage{subcaption}
\usepackage{caption}
\usepackage[margin=1in]{geometry}
\usepackage{pifont}
\usepackage{tikz}
\usepackage{lipsum}
\usepackage[most]{tcolorbox}
\usepackage{lineno}
\usepackage{titletoc}
\usepackage{tocloft}
\usepackage{pgfplots}
\usepackage{pgfplotstable}
\usepackage{makecell}
\usepackage{colortbl}
\usepackage{wrapfig}
\usepackage{amsmath,amsfonts}
\usepackage{algorithmic}
\usepackage{array}
\usepackage{textcomp}
\usepackage{stfloats}
\usepackage{url}
\usepackage{verbatim}
\usepackage{balance}

\def\BibTeX{{\rm B\kern-.05em{\sc i\kern-.025em b}\kern-.08em
    T\kern-.1667em\lower.7ex\hbox{E}\kern-.125emX}}
\usetikzlibrary{positioning}
\usetikzlibrary{backgrounds}
\usetikzlibrary{patterns}
\usetikzlibrary{fit}
\usetikzlibrary{shapes.geometric}
\usetikzlibrary{calc}
\pgfplotsset{compat=1.17}

\title{PALM: Enhanced Generalizability for Local Visuomotor Policies via Perception Alignment}

\author{$\text{Ruiyu Wang}^{*}$, $\text{Zheyu Zhuang}^{*}$, Danica Kragic and Florian T. Pokorny
\thanks{* Equal contribution.}
\thanks{This work was partially supported by the Wallenberg AI, Autonomous Systems and Software Program (WASP) funded by the Knut and Alice Wallenberg Foundation. \textit{Corresponding author: Ruiyu Wang}.}
\thanks{All authors are with the Division of Robotics, Perception and Learning, KTH Royal Institute of Technology, 10044 Stockholm, Sweden (e-mail: \{ruiyuw, zheyuzh, dani, fpokorny\}@kth.se).}}

\begin{document}

\maketitle
\thispagestyle{empty}
\pagestyle{empty}

\begin{abstract}
    Generalizing beyond the training domain in image-based behavior cloning remains challenging. Existing methods address individual axes of generalization, workspace shifts, viewpoint changes, and cross-embodiment transfer, yet they are typically developed in isolation and often rely on complex pipelines.
We introduce PALM (Perception Alignment for Local Manipulation), which leverages the invariance of local action distributions between out-of-distribution (OOD) and demonstrated domains to address these OOD shifts concurrently, without additional input modalities, model changes, or data collection.
PALM modularizes the manipulation policy into coarse global components and a local policy for fine-grained actions.
We reduce the discrepancy between in-domain and OOD inputs at the local policy level by enforcing local visual focus and consistent proprioceptive representation, allowing the policy to retrieve invariant local actions under OOD conditions.
Experiments show that PALM limits OOD performance drops to 8\% in simulation and 24\% in the real world, compared to 45\% and 77\% for baselines. \href{https://github.com/RuiyuWANG/PALM}{{\raisebox{-0.2ex}{\includegraphics[height=1em]{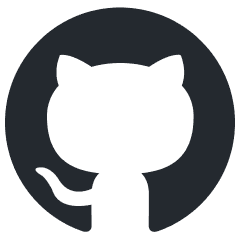}}}}
\end{abstract}


\section{Introduction}
    \IEEEPARstart{V}{isuomotor} policies, which map visual observations and robot proprioceptive states directly to control signals, have become a widely adopted approach for robotic manipulation~\cite{mandlekar_what_2021}.
However, end-to-end policies tend to overfit to the training distributions~\cite{gao2024}, which limits their ability to generalize to out-of-domain scenarios~\cite{pumacay2024colosseum,xie2024decomposing}.
Previous works address this by expanding dataset diversity~\cite{embodimentcollaboration2024openxembodimentroboticlearning,khazatsky2024droid}, which is labor-intensive.

Visual data augmentation provides a data-efficient alternative, via generative models~\cite{yu2023scaling,chen2024semantically} or image superimposition~\cite{zhuang2024enhancing}. 
However, these techniques operate solely in the visual domain and do not account for variations in proprioceptive states (e.g., end-effector (EE) pose), thereby limiting their utility to visual-only shifts such as distractors, lighting, or scene backgrounds.

\begin{figure}[t]
    \centering
    \includegraphics[width=0.8\linewidth]{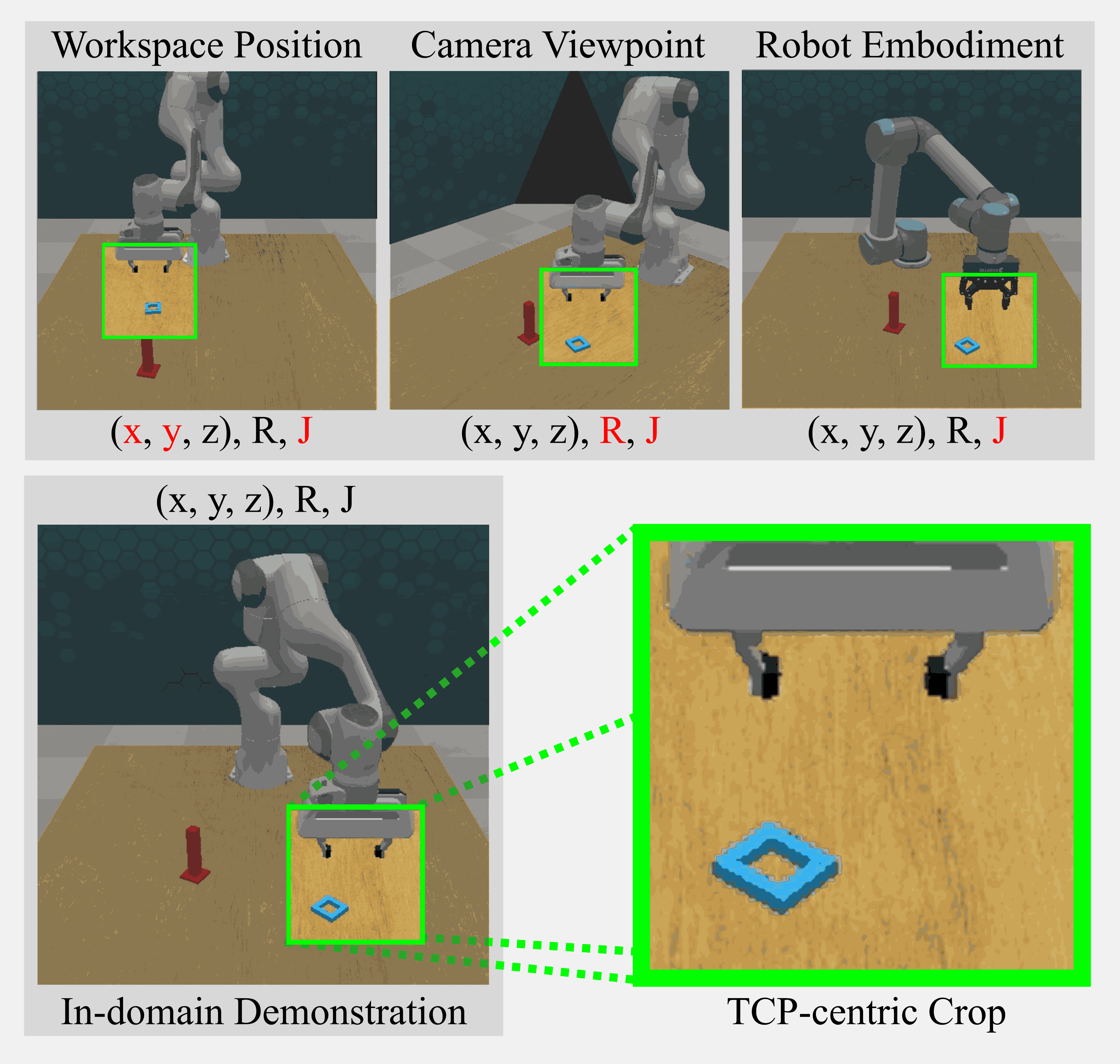}
    \captionof{figure}{\textbf{Input modalities under OOD shift.} 
    Workspace, camera, and embodiment shifts cause significant variations in the third-person image relative to in-domain settings, while changes in the local area near the end-effector remain less pronounced. Corresponding misalignments (in red) also occur in proprioceptive states, e.g., EE $(x, y)$, rotation $\mathbf{R}$, and robot joint angles $\mathbf{J}$.}
    \label{fig:teasor}
    \vspace{-2mm}
\end{figure}

Changes that alter both critical visual features and proprioceptive inputs remain difficult to handle and are typically studied in isolation, for example, robot workspace~\cite{mandlekar2023mimicgen,yin2023spatial}, camera viewpoint~\cite{shang2021self,chen2024roviaug}, and embodiment~\cite{lepert2025shadow,chen2024mirage}.
These approaches often require intensive sampling in simulation~\cite{shang2021self}
or additional input modalities such as eye-in-hand cameras~\cite{hsu2022vision} or voxel-based representations~\cite{wang2024equivariant,zhao2025hierarchical}.
However, paired simulation environments or extra input modalities are not always available in existing datasets~\cite{embodimentcollaboration2024openxembodimentroboticlearning}, and are not universally adopted in practice.
In contrast, third-person cameras are widely used as the only visual input because they offer a global context of both robot–object interactions and subtask transitions, which facilitates long-horizon, multi-stage tasks with a single visual modality~\cite{mandlekar_what_2021,wang_mimicplay_2023}.
These motivate us to propose a \textit{unified} approach that enables visuomotor policies integrated with third-person camera to generalize concurrently across shifts in workspace position, camera viewpoint, and robot embodiment\footnote{For brevity, we refer to these three types of shift collectively as OOD shift unless explicitly stated otherwise.}, without additional input modalities, model modifications, or data collection.

A key yet often overlooked invariance across in-domain and OOD settings lies in the local interactions between the robot and the objects, namely, the local action distribution.
This is particularly illustrative for workspace shift: intuitively, picking up a cube from the left side of the table requires similar motions as picking from the right.
However, trained policies often fail to retrieve these actions due to drastic changes in the observation space, such as robot appearance when the camera is rotated or end-effector positions under workspace shifts, as shown Fig.~\ref{fig:teasor}.
To exploit this local action invariance, we propose Perception Alignment for Local Manipulation (PALM). 
PALM modularizes a manipulation policy into coarse global actions, which move the robot near target objects, and a precise, generalizable local policy trained from demonstrations, aiming for stronger generalization when combined.
The global actions are realized by an analytical global policy (Sec.~\ref{method:localmanipualtion}), and the local policy is enhanced through input-level alignment across two modalities.

\textbf{Visual Alignment.} 
We apply a tool center point (TCP)-centric crop on the third-person images that canonicalizes the visual input on a local region near the end-effector. 
This local visual focus reduces variation due to workspace, viewpoint, and embodiment changes (see Fig.~\ref{fig:teasor}), while preserving essential task information, since physical interaction between robots and objects always occurs locally near the end effector~\cite{johns2021coarse}.

\textbf{Proprioceptive Alignment.} 
We exclude $(x, y)$ coordinates from proprioceptive inputs and represent the end-effector’s rotation relative to the camera frame.
This design effectively decouples the policy from a fixed global frame and yields consistent proprioceptive representation across in-domain and OOD settings.

These alignments explicitly match in-domain and OOD observations, allowing the local policy to retrieve the learned actions under OOD conditions.
Implemented as a pre-processing step, PALM leaves the training pipeline unchanged and supports simultaneous generalization across shifts in workspace, camera, and embodiment, without requiring additional data, modalities, or model changes.
We evaluate PALM on four manipulation tasks in simulation and two tasks on a real robot, showing that it reduces OOD performance drop to 8\%, compared to 45\% for the \textit{best} baseline in simulation and 24\% versus 77\% drop in the real world.

\begin{figure}
    \centering
    \includegraphics[width=1.0\linewidth]{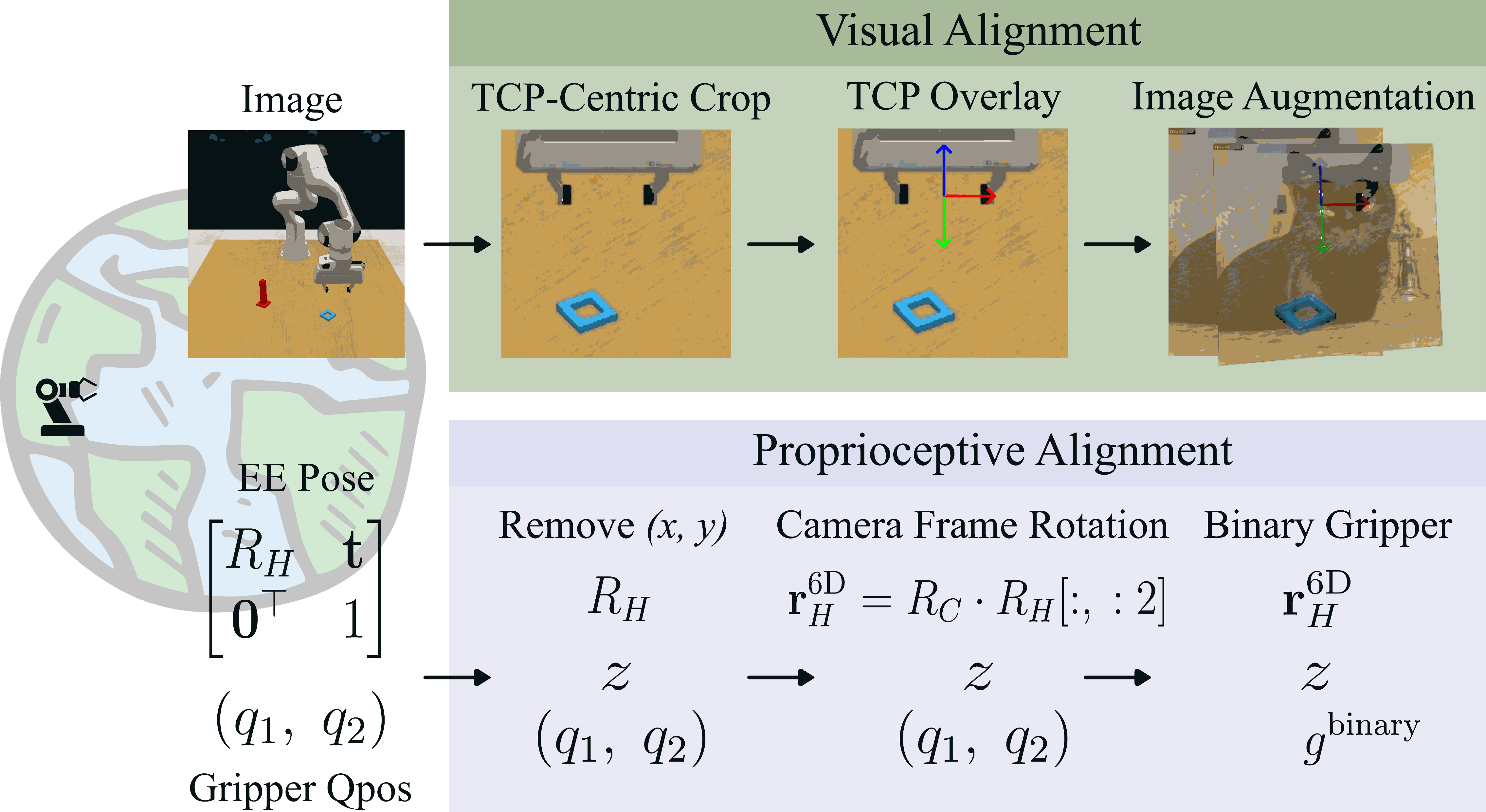}
    \caption{\textbf{Alignments for local policy.} PALM performs visual and proprioceptive alignment on the input as data pre-processing for local manipulation policies.}
    \label{fig:palm_method}
    \vspace{-2mm}
\end{figure}
    
\section{Related Work}
     To the best of our knowledge, there is no method that adequately addresses all three OOD settings; we therefore review the literature in each area separately.

\textbf{Workspace Generalization.}
Prior methods rely on curated data to broaden spatial coverage.
Object-centric approaches~\cite{mandlekar2023mimicgen,xue2025demogen} assume known object poses and generate synthetic data through object-centric transformation of demonstrations. 
MirrorDuo~\cite{mirrorduo2025} lifts this assumption by exploiting reflection symmetry.
Coarse-to-fine manipulation methods~\cite{johns2021coarse,valassakis2022demonstrate} depend on trained controllers to reposition the end-effector near objects and perform state-based replay, but struggle to maintain performance in long-horizon tasks. 
In contrast, PALM does not rely on object pose estimation and supports long-horizon tasks through its modular structure.
    
\textbf{Viewpoint Robustness.}
Robustness to camera motion solely based on 2D image input remains challenging.
Prior work addresses this through view-invariant representation learning~\cite{yuan2024learning,seo2023multi}, but typically requires extensive viewpoint sampling in simulation, hindering real deployment.
Other approaches leverage NeRF-based novel view synthesis for eye-in-hand cameras~\cite{zhou2023nerf,zhang2024diffusion} and third-person views~\cite{chen2024roviaug}, yet require data synthesis and retraining. In contrast, PALM achieves viewpoint robustness without relying on pretrained models or sampling.
    
\textbf{Cross-Embodiment.}
Generalization across robots has been addressed by removing the original robot via segmentation masks~\cite{hu2021know, mirjalili2025augmented}, synthesizing robot appearances with diffusion models~\cite{chen2024mirage,chen2024roviaug}, overlaying robot images~\cite{lepert2025phantom} or aligning internal representations with contrastive learning~\cite{yang2023polybot}.
PALM does not rely on high-precision robot masks or diffusion models, which hinder real-time inference.
    
\section{Methodology}
    To exploit local action invariance, PALM decomposes the manipulation policy into coarse end-effector (EE) actions for approaching target objects and generalizable local policies for fine-grained robot–object interactions.
Demonstration data is used exclusively to train the local policy, where image–proprioception pairs $(I,\,\mathbf{p})$ are preprocessed via visual and proprioceptive alignment (Fig.~\ref{fig:palm_method}). 
The camera extrinsic \( X_C \in \mathbb{R}^{4 \times 4} \) and intrinsic \( K \in \mathbb{R}^{3 \times 3} \) matrices are assumed to be known.

\subsection{Visual Alignment}
\label{method:visual_align}

\begin{figure}[h]
    \centering
    \input{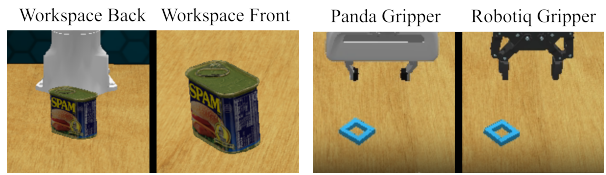}
    \caption{Visual domain shifts in third-person views that cannot be resolved through TCP-centric cropping alone.}
    \label{fig:method_visual_gaps}
    \vspace{-3mm}
\end{figure}

Despite substantial pixel-level variations under OOD shifts, e.g. in background or joint configurations, only a small region near the EE is typically relevant to task progression in many manipulation tasks where no complex contact-rich interaction occurs away from the grasped area.
PALM adopts a tool center point (TCP)-centric crop of the third-person image as its core visual alignment strategy.
However, cropping alone does not fully close the visual gap (Fig.~\ref{fig:method_visual_gaps}).
We therefore introduce TCP Overlay and visual data augmentation as complementary strategies to further mitigate visual discrepancies.

\textbf{TCP-Centric Crop.} 
Given a third-person image \( I \in \mathbb{R}^{h \times w} \), the corresponding EE pose (rotation \( \mathbf{R}_{H} \in \mathbb{R}^{3 \times 3} \) and translation \( \mathbf{t}_{H} \in \mathbb{R}^3 \)) in the world frame and camera matrices $K,X_{C}$, \( \mathbf{t}_{H} \) is projected into the pixel space as:
\begin{equation}
    p_H = \pi\left( K \cdot 
    \begin{bmatrix} \mathbf{R}_C & \mathbf{t}_C \end{bmatrix} 
    \cdot 
    \begin{bmatrix} \mathbf{t}_{H} \\ 1 \end{bmatrix} \right), 
    \pi\left(
        \begin{bmatrix} x \\ y \\ z \end{bmatrix}
    \right) = 
    \begin{bmatrix} x/z \\ y/z \end{bmatrix},
\end{equation}
where \( \mathbf{R}_C\), \( \mathbf{t}_C \) are the rotation and translation components of \( X_C \),  $\pi$ is a projection function defined above.

PALM then performs a fixed size \( (\kappa, \kappa) \) spatial crop on image $I$, centered at \( p_H \in \mathbb{R}^2 \):
\begin{equation}
    I^{\text{crop}} = \textit{CenterCrop}(I, p_H, \kappa).
\end{equation}

\begin{figure}[h]
    \centering
    \includegraphics[width=1.0\linewidth]{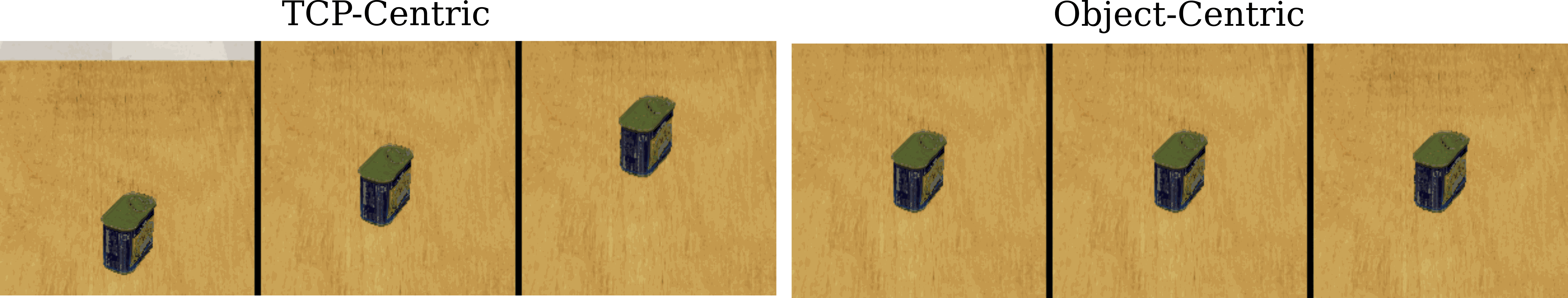}
    \captionof{figure}{TCP-centric vs. object-centric cropping at time steps 1, 18, and 50 on \textit{Lift Spam Invisible}.
    TCP cropping carries the robot's movement across frames.}
    \label{fig:object_centric_crop}
    \vspace{-3mm}
\end{figure}

TCP-centric cropping stabilizes OOD visual variation in the full third-person camera views and maps the end-effector to a consistent pixel location.
As shown in Fig.~\ref{fig:object_centric_crop}, it implicitly projects end-effector motion into image space. 
This property distinguishes PALM from object-centric methods and approaches that simply mask irrelevant content. 
This distinction is further confirmed by a conceptual experiment in Tab.~\ref{tab:object_centric_crop}.

\textbf{TCP Overlay.} 
An overlay of the pixel space projection of three EE rotation axes, each in different colors, is applied on top of \( I^{\text{crop}} \) at \( p_{H} \), see Fig.~\ref{fig:palm_method}.
This overlay requires no additional inputs yet provides a consistent visual feature that substitutes for the specific gripper appearance, improving cross-embodiment transfer (Tab.~\ref{tab:exp_main_ablation}).

\textbf{Data Augmentation.}
To mitigate remaining visual shifts across objects and embodiments, as shown in Fig.~\ref{fig:method_visual_gaps}, PALM applies two image augmentation techniques: Random Overlay~\cite{zhuang2024enhancing}, which superimposes irrelevant images on image inputs to encourage policies' focus on task-relevant regions, and a mild perspective transformation~\cite{torchvision2016} to introduce viewpoint diversity (Fig.~\ref{fig:palm_method}).
Such augmentations prompt invariant feature learning, further improving cross-embodiment generalization (see Tab.~\ref{tab:exp_main_ablation}).
They also improve policy robustness to cluttered scenes, lighting changes, and distractors, especially in real-world settings.

\subsection{Proprioceptive Alignment}
\label{method:prop_align}
The choices of proprioceptive states span a collection of EE poses, joint angles, gripper positions, commonly expressed in robot base frame.
Their role in generalization, to our best knowledge, has not been systematically studied, yet we find that achieving consistency across domains requires a deliberate formulation.
For example, EE $(x,y)$ coordinates may fall outside the training distribution for a given workspace, and joint angles may fail to transfer across embodiments.
PALM’s proprioceptive alignment follows two core principles: (1) selecting states that are decoupled from a fixed global frame to support spatial shifts, and (2) representing them in a form invariant to camera viewpoint and embodiment.
Specifically, PALM uses the $\mathrm{SE}(3)$ EE pose and gripper joint positions as the proprioceptive modality $\mathbf{p}$, represented as:

\textbf{Height-Only Translation.} 
PALM excludes the $(x, y)$ coordinates from translation inputs. 
This decision is grounded in the observation that manipulation datasets generally provide adequate coverage along the $z$-axis, whereas achieving comprehensive sampling across the $(x, y)$ plane, particularly in multi-stage settings, requires prohibitively expensive data collection.
$(x, y)$  coordinates furthermore couple the input to a fixed global frame, making them incompatible with TCP-centric cropping, as visually similar images may correspond to different $(x, y)$ values. 
The thus argue that this spatial information can instead be inferred directly from the image data.

\textbf{Camera Frame Rotation.}
Although the demonstration data covers a wide range of rotations, the initial pose of the robot remains relatively constrained across trajectories and becomes misaligned under camera rotation when expressed in the world frame.
PALM resolves this by expressing the EE rotation $R_H$ in the camera coordinate frame, $^{C}{R}_{H} = R_C^\top \cdot R_H$, and adopting a 6D rotation representation~\cite{hempel20226d}, $\mathbf{r}^{6D}_H = \phi (^{C}{R}_{H})$ where $\phi : \mathrm{SO}(3) \rightarrow \mathbb{R}^6$.
This ensures consistent rotation states under different camera orientations.

\textbf{Binary Gripper State.}
PALM expresses the gripper states as a binary number of close (0) and open (1). 
This excludes the difference in gripper poses, due to different mechanical structures or gripper sizes, enabling polices to transfer between a variety of grippers.

This final proprioception state is then expressed as:
\begin{equation}
    \mathbf{p}^{\text{align}} = (\mathbf{r}^{\text{6D}}_{H},\;z,\; g^{\text{binary}}).
    \vspace{-3mm}
\end{equation}

\begin{figure*}[th]
    \centering
    \includegraphics[width=0.95\linewidth]{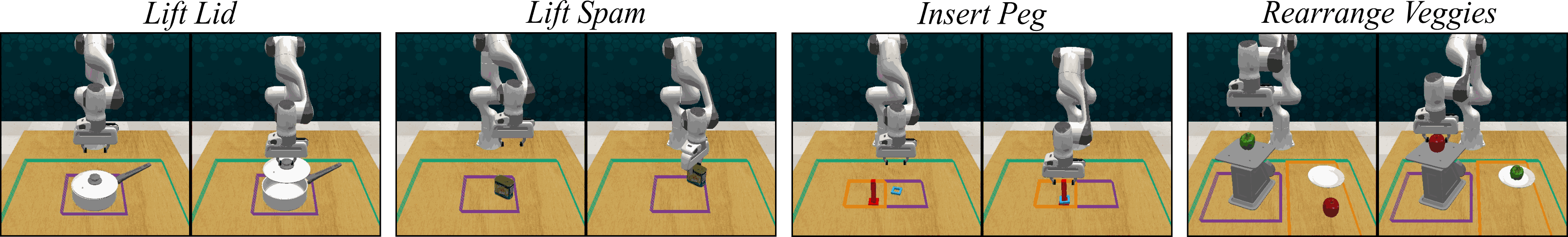}
    \caption{\textbf{Tasks and OOD setups.} Four simulation tasks with testing workspace ranges shown in green and training ranges in other colors. Testing domains for a selected camera viewpoint and cross-embodiment are shown in Fig.~\ref{fig:teasor}.}
    \label{fig:task_sim}
    \vspace{-3mm}
\end{figure*}

\subsection{Beyond Local Manipulation}
\label{method:localmanipualtion}
While visual and proprioceptive alignment are effective for local manipulation, long-horizon multi-stage tasks require the policy to infer task progression. 
Such information may not be present in cropped images, and actions for reaching target objects at arbitrary workspace locations are missing from most legacy datasets. 
In contrast, full third-person images preserve high-level task information, and reaching primarily involves coarse motion that does not need to be learned from demonstration, enabling the reaching phase to be addressed analytically. 
PALM therefore modularizes the policy into a coarse analytical global policy that performs spatial reasoning over full third-person images and a fine-grained, generalizable local policy.

As a feasibility demonstration, we use a simple analytical global policy that estimates target object positions $(x_O, y_O)$ by projecting the center of object segmentation from pixel space to the robot frame. 
During evaluation, the global policy first repositions the end-effector $(x, y)$ to a randomly sampled position centered at $(x_O, y_O)$ using position-based control (no rotation involved), after which the local policy infers actions from the aligned inputs. Across multi-stage tasks in simulation (\textit{Insert Peg}, \textit{Rearrange Veggies}) and the real world (\textit{Drawer}, \textit{Stack}), this modular design improves generalization even with a simple global policy. Global policies, such as finetuned foundation models~\cite{kim24openvla}, may yield further gains, which we leave for future work.
    
\section{Experiments}
    \begin{table*}[t]
\centering
\begin{tabular}{ccccc|cccc|c}
 & \multicolumn{4}{c}{\text{Baselines}} & \multicolumn{4}{c}{\text{Ablations}} & \\
\toprule
\text{Testing Domain} & {MirrorDuo} & {RoVi-Aug} & {ARRO} & BC & P $\ominus$ \textit{Crop} & {P $\ominus$ \textit{Overlay}} & P $\ominus$ \textit{Aug} & P $\ominus$ \textit{Prop} & PALM \\
\midrule
\textit{In-domain} ($\uparrow$) & {0.47} & {0.26} & {0.41} & 0.52 & 0.35 & \textbf{0.73} & \textbf{0.73} & 0.70 & 0.72 \\
\midrule
\multicolumn{10}{c}{\textit{Out-of-domain} ($\downarrow$)} \\
\midrule
\textit{Workspace} & {0.33} & {0.17} & {0.25} & 0.40 & 0.23 & {0.08} & 0.08 & 0.35 & \textbf{-0.02} \\
\textit{Viewpoint} & {0.52} & {0.36} & {0.55} & 0.61 & 0.30 & {0.11} & \textbf{0.05} & 0.32 & 0.06 \\
\textit{Embodiment} & {0.60} & {0.41} & {0.32} & 0.79 & 0.65 & {0.47} & 0.23 & 0.15 & \textbf{0.10} \\
\textit{Workspace + Viewpoint} & {0.61} & {0.45} & {0.81} & 0.65 & 0.35 & {0.04} & 0.06 & 0.55 & \textbf{0.00} \\
\textit{Workspace + Embodiment} & {0.67} & {0.55} & {0.48} & 0.78 & 0.74 & {0.45} & 0.21 & 0.53 & \textbf{0.05} \\
\textit{Viewpoint + Embodiment} & {0.70} & {0.72} & {0.60} & 0.89 & 0.73 & {0.44} & 0.34 & 0.38 & \textbf{0.21} \\
\textit{All} & {0.65} & {0.49} & {0.79} & 0.83 & 0.66 & {0.44} & 0.21 & 0.57 & \textbf{0.18} \\
\midrule
\textit{OOD average ($\downarrow$)} & {0.58} & {0.45} & {0.54} & 0.71 & 0.52 & {0.29} & 0.17 & 0.41 & \textbf{0.08} \\
\bottomrule
\end{tabular}

\caption{\textbf{Main experiments and ablations.} 
In-domain success rate ($\uparrow$) and OOD normalized degradation ($\downarrow$) for three baselines, vanilla BC (BC) and ablations including: PALM without without TCP-centric crop (P $\ominus$~{\textit{Crop}}), without TCP Overlay (P $\ominus$~{\textit{Overlay}}), without data augmentation (P $\ominus$~{\textit{Aug}}), without proprioception alignment (P $\ominus$~{\textit{Prop}}), and the full PALM pipeline, averaged over four simulated tasks.}
\label{tab:exp_main_ablation}
\vspace{-3mm}
\end{table*}
\label{exp:main}
\textbf{Simulation Task Setup.}
We evaluate PALM on four RLBench tasks~\cite{james2020rlbench} selected to cover a range of difficulty and motion patterns, from simple tasks such as \textit{Lift Lid}, to high-precision tasks like \textit{Lift Spam} and \textit{Insert Peg}, and a long-horizon, cluttered-scene task, \textit{Rearrange Veggies}.
Training datasets contain 300 demonstrations for the two multi-stage tasks and 200 for the rest, collected with objects sampled in constrained ranges, a fixed front camera, and a Franka Panda robot.
During testing, objects are sampled uniformly within an expanded workspace, see Fig.~\ref{fig:task_sim}. 
Viewpoint shifts are introduced by rotating the camera around the workspace center’s z-axis by a random angle within $[-30^\circ, 30^\circ]$.  
Cross-embodiment is performed by testing the policies on a UR5 robot with a Robotiq gripper, see Fig.~\ref{fig:teasor}.  

\begin{figure}[h]
    \centering
    \includegraphics[width=0.95\linewidth]{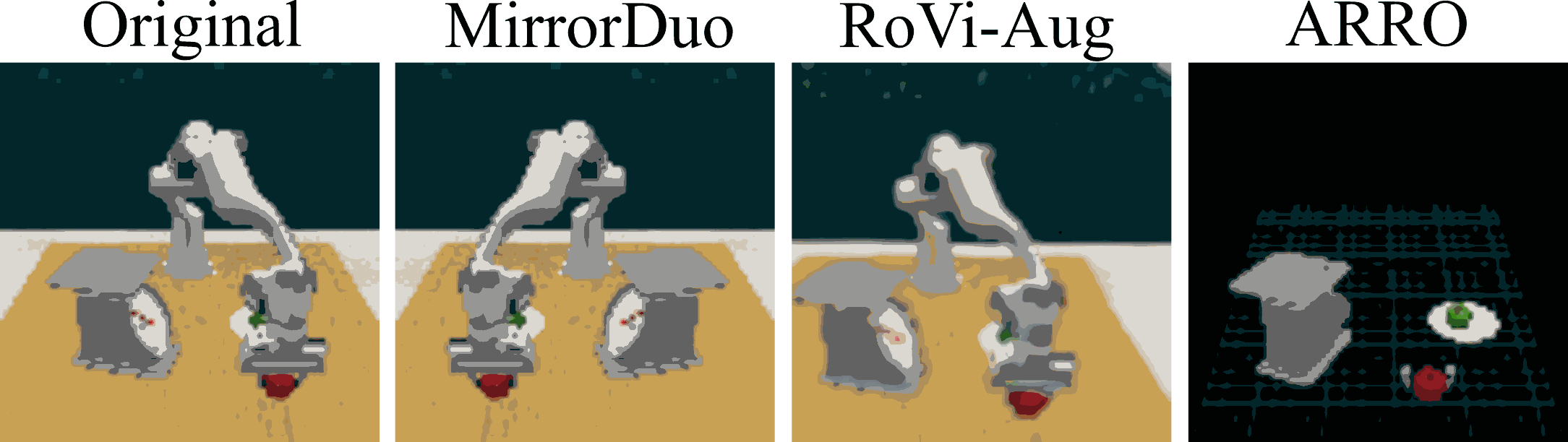}
    \caption{Example of the effect of image augmentation or pre-processing adopted in the baselines.}
    \label{fig:baseline_example}
    \vspace{-2mm}
\end{figure}

\textbf{Baseline and Evaluation Protocol.}
For evaluation, PALM is integrated with a standard behavior cloning (BC) model~\cite{NIPS1988812b4ba2}. 
Since no prior method simultaneously addresses generalization across three domains, we compare against one representative baseline from each direction, along with PALM variants. Specifically:
\begin{itemize}
    \item 
    MirrorDuo~\cite{mirrorduo2025}, which augments whole demonstrations via mirroring to enable workspace transfer from one-sided data. We evaluate its ability in improving workspace generalization.
    
    \item 
    RoVi-Aug~\cite{chen2024roviaug}, which synthesizes novel viewpoints and robot embodiments using novel view synthesis and diffusion inpainting. We compare only its viewpoint augmentation (see Fig.~\ref{fig:baseline_example}) due to its lack of cross-embodiment support for simulation datasets.
    
    \item 
    ARRO~\cite{mirjalili2025augmented}, which segments the gripper and objects and pastes them onto a black background with grid pattern. We use this as a segmentation-based baseline for cross-embodiment transfer.
\end{itemize}

Models are trained for 600 epochs, and performance is measured using the best of the top three in-domain checkpoints, saved every 50 epochs and evaluated on 50 scenes each. 
We report the average success rate $\eta$ and the OOD normalized degradation, $1 - \eta^{\text{OOD}}/\eta^{\text{In-domain}}$. 
All comparisons focus on the local policy, while the global policy is identically applied across methods.

\textbf{Implementation Details.}
All models are initialized with a ResNet-18~\cite{he2016deep} ImageNet pretrained visual encoder, followed by a two-layer MLP.
Raw visual inputs of size $512 \times 512$ are resized to $84 \times 84$ for model input, either directly for baselines and vanilla BC or after TCP-centric cropping for PALM.
The implementation of Random Overlay follows~\cite{zhuang2024enhancing}.
We use a relative action mode for EE pose, as absolute action is attached to a fixed frame and does not generalize beyond the demonstrated workspace coverage.
MirrorDuo and RoVi-Aug follow their public code releases. 
Since ARRO is not open-source, we reimplement it with ground-truth segmentation from the simulator (see Fig.~\ref{fig:baseline_example}).

\subsection{The Effectiveness of PALM: A Synergy of Visual and Proprioceptive Alignment}
\label{exp:main_res_and_ablation}
\begin{figure}[t]
    \centering
    \includegraphics[width=1.0\linewidth]{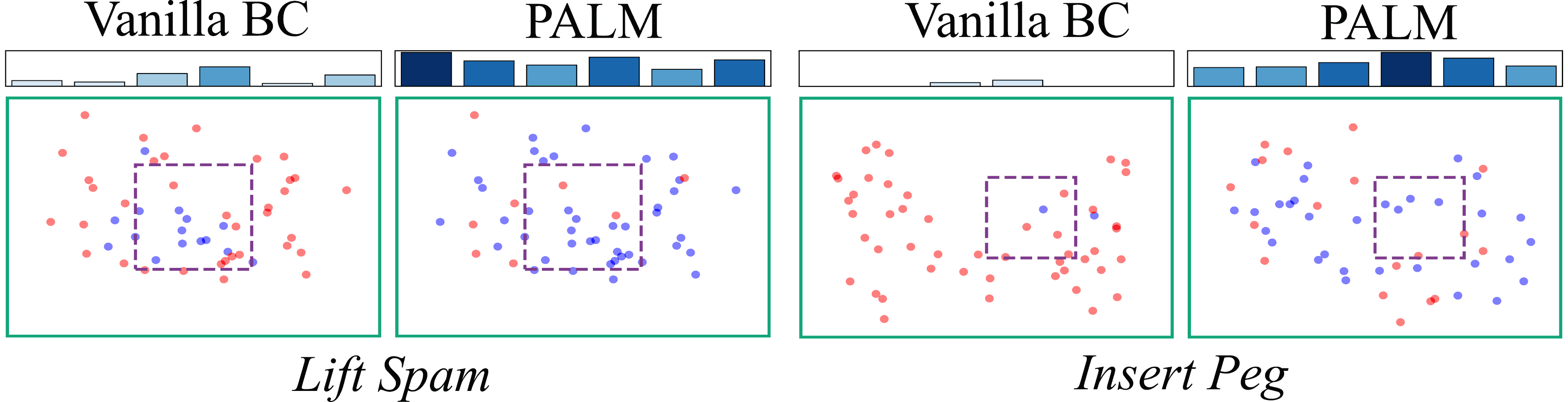}
    \caption{\textbf{Sample Distribution.} For each cell, Top: success rate bar plot for camera shifts across six bins in $[-30^\circ, 30^\circ]$; Bottom: sampled object $(x, y)$ positions colored by outcome (blue: success, red: failure); green/purple bounds mark test/train workspaces.}
    \label{fig:sample_distribution}
    \vspace{-3mm}
\end{figure}

We evaluate PALM under various shifts in workspace position, camera viewpoint, embodiment and their combinations, as shown in Tab.~\ref{tab:exp_main_ablation}.  
PALM markedly improves policy robustness for third-person datasets, reducing the average OOD normalized degradation to 8\%, compared to 71\% for vanilla BC.  
Fig.~\ref{fig:sample_distribution} further illustrates that vanilla BC fails to generalize when objects are sampled outside the training workspace range and can handle only minor camera rotations (within 10°), whereas PALM preserves performance in OOD scenarios.
Three baselines improve OOD performance in their respective domains relative to vanilla BC, but none match PALM in either single-shift or combined settings. 
RoVi-Aug exhibits lower in-domain performance, likely due to reconstruction error introduced by the diffusion model.
PALM can also be applied on top of these baselines.

To isolate the contributions of PALM’s components, we evaluate two ablations: P~$\ominus$~{\textit{Prop}} (no proprioceptive alignment) and P~$\ominus$~{\textit{Crop}} (no crop). 
Tab.~\ref{tab:exp_main_ablation} shows that removing either significantly harms generalization, suggesting that aligning \textit{both} modalities is essential.
TCP-centric crop is particularly important for cross-embodiment, whereas proprioceptive alignment is crucial under workspace and camera shifts.  
Removing TCP Overlay (P $\ominus$~{\textit{Overlay}}) and data augmentation  (P $\ominus$~{\textit{Aug}}) has minimal effect on workspace and viewpoint but yields 37\% and 13\% degradation on embodiment.
This confirms cropping as the primary driver of visual alignment when scene appearance varies largely and shows that TCP Overlay supplies embodiment-consistent cues that masking alone, as in ARRO, cannot provide.

\subsection{Ablations and Other Key Findings}
\label{exp:other_ablatons}
We test PALM's capability through experiments designed to answer the following questions.

\begin{figure}[t]
    \centering
    \includegraphics[width=0.82\linewidth]{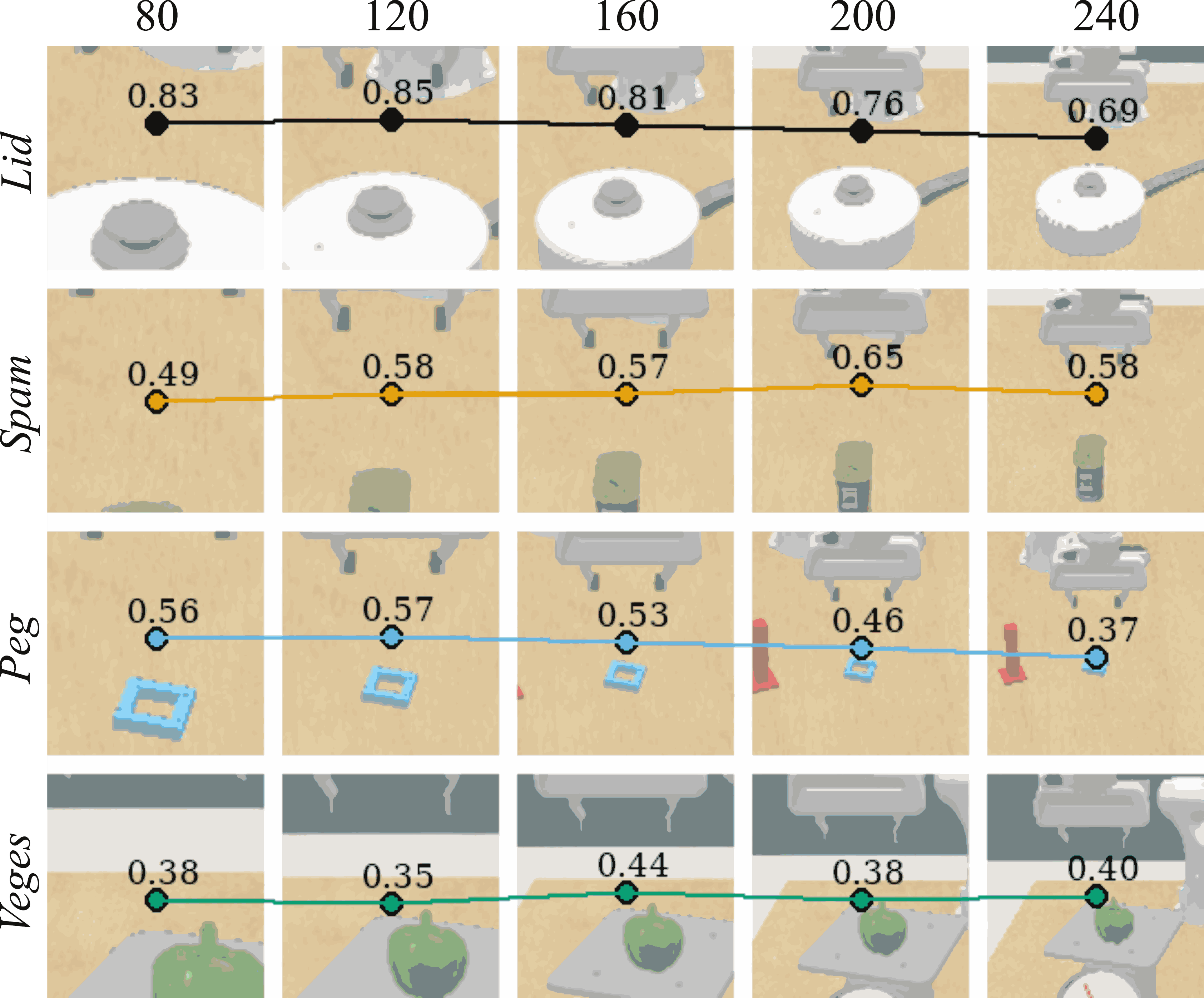}
    \caption{\textbf{Crop size ablation.} OOD success rates of PALM with cropped images of size 80, 120, 160, 200, and 240.}
    \label{fig:crop_size}
    \vspace{-3mm}
\end{figure}

\noindent \textbf{Q1}: How does PALM respond to crop size and calibration accuracy?

TCP-centric cropping treats crop size as a hyperparameter controlling how locally the policy attends. 
We evaluate PALM over a range of crop sizes, and as shown in Fig.~\ref{fig:crop_size}, OOD performance remains stable for sizes between 80 and 200. 
Intuitively, optimal performance occurs when the crop covers both the gripper and key object regions.
To test PALM’s sensitivity to camera calibration, we jitter the EE projection $p_H$ by $\pm 5$ pixels when cropping out of a $512 \times 512$ image and add Gaussian noise with scale 0.005 to the proprioceptive states during training and evaluation.
In practice, calibration re-projection error is usually under 1 pixel~\cite{boby2016single}. 
Under such perturbation, PALM achieves an in-domain success rate of 0.69, with OOD drops of $-0.11$, $-0.07$, and $0.25$ on workspace, viewpoint, and embodiment, averaged over four simulated tasks. 
This indicates PALM remains robust under reasonably large calibration error, with embodiment performance reduced mainly due to jitter-induced noise in TCP Overlay.

\noindent \textbf{Q2}: How does PALM adapt to extra visual modalities?

Additional visual modalities, such as an eye-in-hand camera, provide an equivariant perspective with respect to workspace and camera, and have been employed in earlier work to improve OOD generalization~\cite{hsu2022vision}. 
We test PALM with such a modality. 
Tab.~\ref{tab:res_with_in_hand} shows that it boosts in-domain performance and reduces OOD degradation to 15\% for BC and 4\% for PALM. 
Although this narrows the advantage of PALM over the baseline, adding modalities serves as a partial remedy rather than addressing the core issue of generalizing from third-person views. 
PALM tackles this problem directly, maintaining superiority over BC in all OOD cases while remaining compatible with additional modality inputs.

\begin{table}[th]
    \centering
    \begin{tabular}{ccc}
\toprule
 & BC &  PALM \\
\midrule
\textit{In-domain} ($\uparrow$)  & \textbf{0.89} & 0.88 \\
\midrule
\multicolumn{3}{c}{\textit{Out-of-domain} ($\downarrow$)} \\
\midrule
\textit{Workspace} & 0.03 & \textbf{-0.06} \\
\textit{Viewpoint} & 0.09 & \textbf{-0.01} \\
\textit{Embodiment} & 0.26 & \textbf{ 0.11} \\
\textit{Workspace + Viewpoint} & 0.09 & \textbf{-0.02} \\
\textit{Workspace + Embodiment} & 0.09 &  \textbf{0.08} \\
\textit{Viewpoint + Embodiment} & 0.26 &  \textbf{0.12} \\
\textit{All} & 0.26 &  \textbf{0.06} \\
\midrule
\textit{OOD average ($\downarrow$)}  & 0.15 &  \textbf{0.04} \\
\bottomrule
\end{tabular}
    \caption{\textbf{Additional visual modality.} Performance of vanilla BC and PALM with both third-person and eye-in-hand as visual input modalities.}
    \label{tab:res_with_in_hand}
     \vspace{-3mm}
\end{table}

\noindent \textbf{Q3}: Is the benefit of cropping driven by resolution?

Cropping effectively increases image resolution after resizing and lowers the error-per-pixel (EPP) when mapping pixel-level changes to physical distances in the workspace. 
To isolate the effect of resolution, we train vanilla BC with an input size of $268 \times 268$, matching PALM’s EPP. 
Experiment results show that the in-domain success rate averaged over 4 tasks rises to 73\%, compared with 53\% for BC at 84 resolution.
However, OOD degradation worsens to 81\% versus 71\%. 
This indicates that the benefit of cropping for OOD generalization does not stem from increased resolution.

\noindent \textbf{Q4}: What are the benefits vs. object-centric cropping? 

Local cropping can be achieved by centering either on the TCP or on the target object. 
While these crops often overlap for local manipulation, TCP-centric cropping provides benefits beyond filtering irrelevant content. 
It fixes the robot TCP at a specific image location, implicitly encoding end-effector motion in pixel space. 
To test this, we introduce \textit{Lift Spam Invisible}, a variant of \textit{Lift Spam} where the robot is hidden from view during both training and evaluation. 
As shown in Fig.~\ref{fig:object_centric_crop}, with this design, object-centric cropping cannot recover spatial relations between the robot and objects from images alone, whereas TCP-centric cropping encodes them implicitly.
Tab.~\ref{tab:object_centric_crop} shows that TCP-centric cropping outperforms both object-centric and no-crop policies on \textit{Lift Spam Invisible}. 
When proprioception is limited to $z$, it still preserves performance, while the others degrade further, demonstrating its motion-encoding property.

\begin{table}[h]
    \centering
    \centering
\begin{tabular}{ccc}
\toprule
\text{Crop Type} & \multicolumn{2}{c}{\text{Proprioception}} \\[-0.4ex]
\cmidrule{2-3}
& {$z$} & {$xyz$}  \\[-0.4ex]
\midrule
\textit{No Crop} & 0.18 & 0.40 \\
\textit{Object-centric Crop} & 0.19 & 0.33 \\
\textit{TCP-centric C`rop} & \textbf{0.54} & \textbf{0.58} \\
\bottomrule
\end{tabular}
    \captionof{table}{In-domain success rate $(\uparrow)$ on \textit{Lift Spam Invisible} for three cropping strategies, and proprioceptive input of $\mathbf{r}^{\text{6D}}_{\text{tcp}}$ combined with either $z$ or $xyz$.}
    \label{tab:object_centric_crop}
    \vspace{-2mm}
\end{table}

\noindent \textbf{Q5}: How does PALM perform in the real world?

\begin{figure}[t] 
    \centering
    \begin{subfigure}[t]{0.36\linewidth} 
        \centering         
        \includegraphics[width=\linewidth]{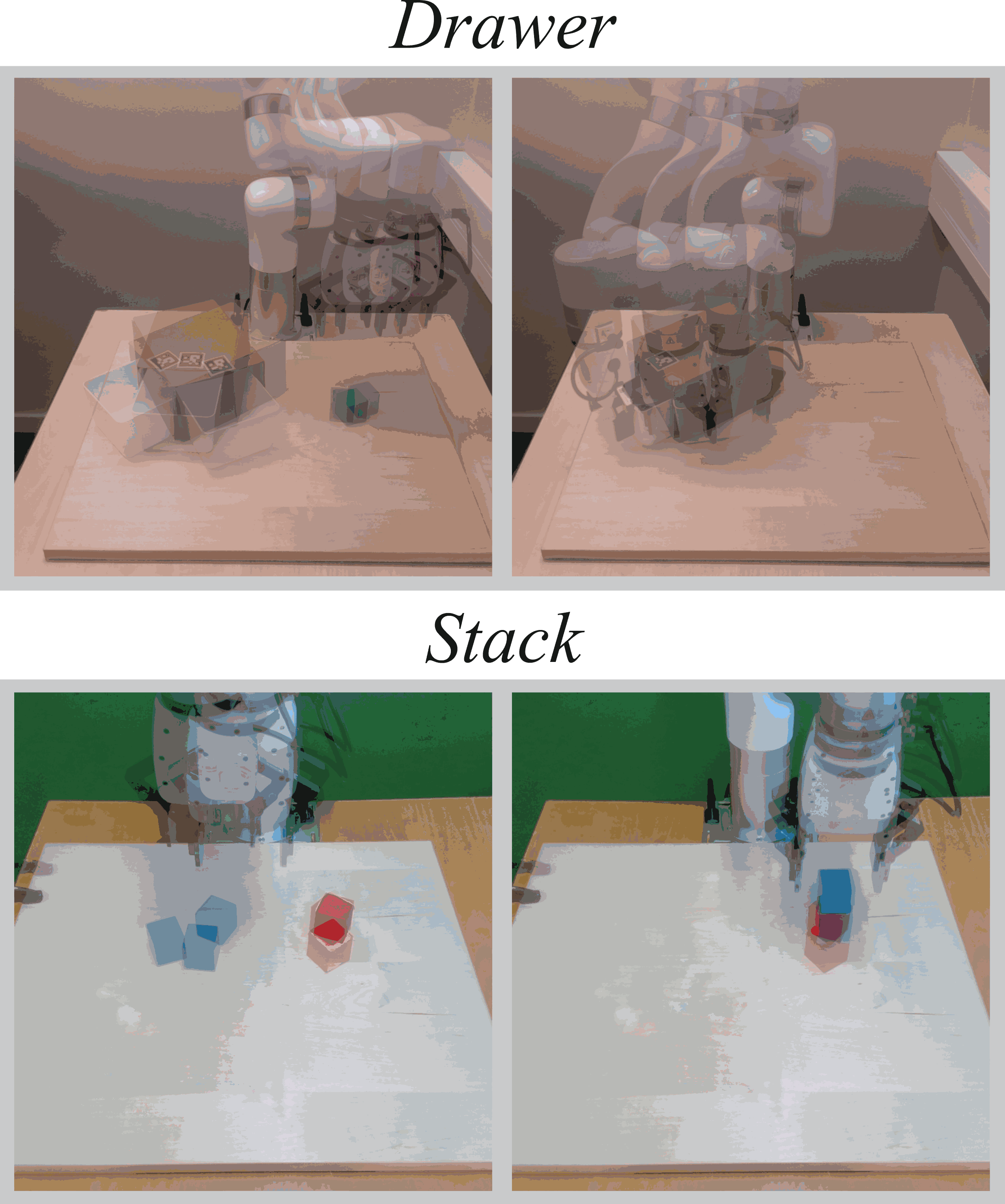} \caption{\small Task distributions.} 
    \label{fig:real_task} 
    \end{subfigure} 
    \hfill 
    \begin{subfigure}[t]{0.62\linewidth} 
        \centering 
        \includegraphics[width=\linewidth]{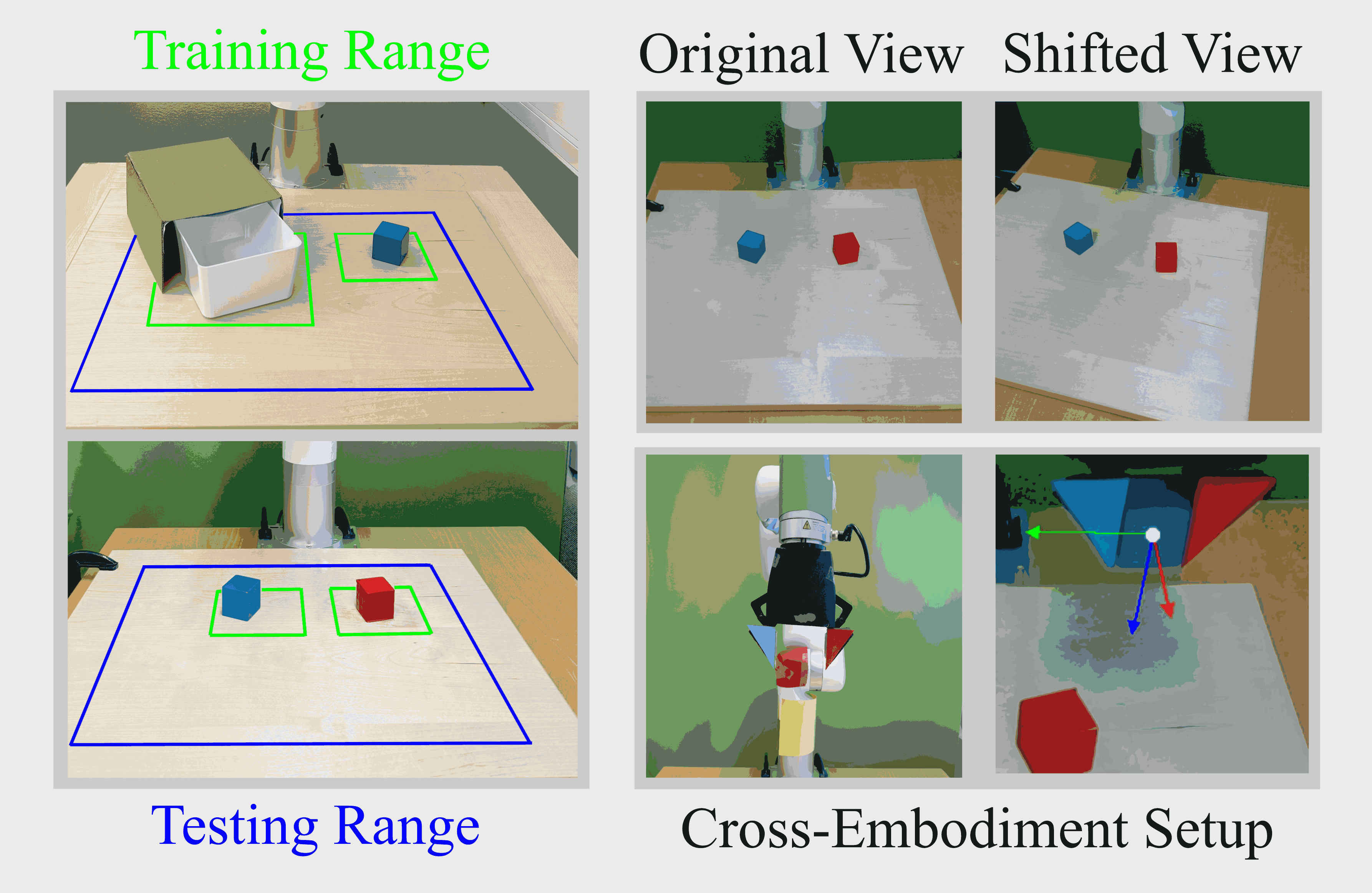} \caption{\small Experiment and OOD setups.} \label{fig:real_setup} 
    \end{subfigure} 
    \caption{\textbf{Real-world experiment tasks and OOD setups.} Viewpoint generalization tested with the camera rotated 15° with respect to the robot base. Workspace is tested with objects randomly sampled in the blue range in (b).} \label{fig:real_exp} \vspace{-3mm} 
\end{figure}

We further evaluate PALM on a real UFactory xArm 7 robot using a single third-person RealSense D435i camera on two multi-stage tasks (Fig.~\ref{fig:real_exp}):
(1) \textit{Drawer}, picking and placing a blue cube into a drawer and closing it, a task characterized by partial occlusions and a relatively long task horizon, instead of only focussing on a short pick-and-place action and  
(2) \textit{Stack}, stacking a blue cube on a red cube, demanding high precision.
Data collection and training follow the protocol in Sec.~\ref{exp:main}, with OOD settings shown in Fig.~\ref{fig:real_exp}.
For each task, we collect 150 demonstrations and evaluate the best checkpoint (lowest validation loss) out of 1000 epochs.

As shown in Tab.~\ref{tab:real_exp}, even with Random Overlay, the BC policy undergoes a drastic performance drop of 77\% in OOD settings, while PALM maintains performance. 
The closer real-world camera setup amplifies visual differences, slightly affecting PALM’s robustness under workspace and camera shifts. 
Due to hardware limits, we approximate cross-embodiment by applying strong visual perturbations (e.g., gripper masking with object-colored textures), under which PALM sustains near-perfect performance.
Both PALM and BC generalize better on \textit{Drawer} than on \textit{Stack}, as the former depends less on precise visual cues and benefits from a larger training range of object placements, although it poses greater in-domain difficulty due to the occlusion.
\textit{Stack} also demands high placement precision of the blue cube, causing BC to completely fail in all OOD settings.
These results demonstrate PALM’s ability to generalize to occluded, long-horizon, and non-pick-and-place settings.

\section{Limitation and Future Work}
    \noindent\textbf{Local crop and multi-stage tasks.}
Our results show that local visual focus, achieved through fixed-size cropping, enhances generalization. 
However, a predefined crop size can be sub-optimal during task progression and requires additional handling for multi-stage scenarios. 
We demonstrate that PALM generally performs best when both the gripper and key object parts are fully captured in the cropped images, and a modular design, separating coarse global actions from a generalizable local policy, offers a promising solution. 
These results point to adaptive cropping, where crop size is dynamically adjusted and learned during task execution, as a natural next step for PALM. 
We consider this an important yet non-trivial direction for future work.

\noindent\textbf{Camera calibration.}
PALM is designed to improve policy generalizability for settings where a fixed and calibrated third-person camera is involved. 
A failure case of the method could occur when calibration is highly inaccurate or not available, e.g. in legacy datasets. 
However, calibration for standalone cameras is a standard practice for real systems and is available in widely used datasets such as DROID~\cite{khazatsky2024droid} and ManiSkill~\cite{gu2023maniskill2}. 
We demonstrate that PALM is tolerant to reasonable calibration error and performs reliably in the real world.
Our results also show that calibration offers a practical trade-off between generalization and data collection effort, making it a reasonable choice for future pipelines. 

\noindent\textbf{Non-tabletop manipulation.}
In this work, workspace generalization is restricted to task workspace translations in the $xy$-plane of the world frame.
If the workspace is positioned off the $xy$-plane, local action invariance fails without further explicit transformation on the state-action pairs, and the proposed method is therefore expected to fail.
This formulation, along with a fixed third-person camera, is commonly adopted in tabletop manipulation.
Settings with moving cameras, e.g., mobile manipulation, are beyond the scope of this work.

\begin{table}[t]
    \centering
\begin{tabular}{ccccc}
\toprule
& \multicolumn{2}{c}{\textit{Drawer}} & \multicolumn{2}{c}{\textit{Stack}}\\[-0.4ex]
\cmidrule(lr){2-3} \cmidrule(lr){4-5}
& BC & PALM & BC & PALM \\[-0.4ex]
\midrule
\textit{In-domain} & 0.65 & \textbf{0.85} & 0.65 & \textbf{1.00} \\
\textit{Workspace} & 0.20 & \textbf{0.60} & 0.00 & \textbf{0.60} \\
\textit{Viewpoint} & 0.20 & \textbf{0.65} & 0.00 & \textbf{0.50} \\
\textit{Embodiment} & 0.50 & \textbf{0.85} & 0.00 & \textbf{1.00} \\
\bottomrule
\end{tabular}
        \captionof{table}{Real experiment success rates $(\uparrow)$ for BC with Random Overlay and PALM. Each data point is averaged over 20 evaluation trials.}
        \label{tab:real_exp}
        \vspace{-3mm}
\end{table}

\section{Conclusion}
    In this work, we identified the invariance of local action distributions between demonstration data and settings with shifts in workspace position, camera viewpoint, and embodiment.
To leverage this property, we introduced PALM (Perception Alignment for Local Manipulation), which modularizes a policy into coarse global motions and a generalizable local policy with aligns inputs via tool-center-point–centric cropping on the third-person image and structured proprioceptive state selection and representation. 
PALM retrieves in-domain actions under OOD scenarios and integrates seamlessly into existing frameworks as a preprocessing step, requiring no additional modalities, model changes, or data collection. 
It is compatible with standard third-person camera data and extends naturally to multi-stage tasks or multi-modality setups. 
Our results indicate that this modular design and local visual focus is a promising direction for generalizable visuomotor policy learning and highlight the importance of carefully selected proprioceptive states for OOD generalization.

\bibliographystyle{IEEEtran}
\bibliography{refs}
\end{document}